\documentclass[proceedings]{ccn}  % options: submission (default), preprint, proceedings, extended-abstract
\usepackage{comment}\usepackage[style=apa,backend=biber]{biblatex}
\DeclareLanguageMapping{american}{american-apa}
\usepackage{float}

\ccndoi{10.32470/05a5frg1}  % required; assigned by CCN

\usepackage{placeins}  % in preamble

\title{Independent-Component-Based Encoding Models of Brain Activity During Story Comprehension}

\author{%
  Kamya Hari\affmark{1} \And
  Taha Binhuraib\affmark{2} \And
  Jin Li\affmark{2} \And
  Cory Shain\affmark{3} \AND
  Anna A. Ivanova\affmark{2} %
}
\affiliation{1}{School of Electrical and Computer Engineering, Georgia Institute of Technology}
\affiliation{2}{School of Psychological and Brain Sciences, Georgia Institute of Technology}
\affiliation{3}{Department of Linguistics, Stanford University}
\emails{khari8, a.ivanova@gatech.edu}

\begin{document}

\maketitle

\begin{abstract}
Encoding models provide a powerful framework for linking continuous stimulus features to neural activity; however, traditional voxelwise approaches are limited by measurement noise, inter-subject variability, and redundancy arising from spatially correlated voxels encoding overlapping neural signals. Here, we propose an independent component (IC)-based encoding framework that dissociates stimulus-driven and noise-driven signals in fMRI data. We decompose continuous fMRI data from naturalistic story listening into ICs using one subset of the data, and train encoding models on independent data to predict IC time series from large language model representations of linguistic input. Across subjects, a subset of ICs exhibited consistently high predictivity. These ICs were spatially and temporally consistent across subjects and included brain networks known to respond during story listening (auditory and language). Auditory component time series were strongly correlated with acoustic stimulus features, highlighting the interpretability of identified component time series. Components identified as noise or motion-related artifacts by ICA-AROMA showed uniformly poor predictive performance, confirming that highly predicted components reflect genuine stimulus-related neural signals rather than confounds. Overall, IC-based encoding models enable analyses at the level of functional networks, accommodating the variability in network locations across individuals and providing interpretable results that are easy to compare across subjects. Code provided at: \url{https://github.com/kamyahari/IC-Encoding-Models.git}
\end{abstract}

\section{Introduction}

Encoding models (EMs) have become a central tool for relating complex, naturalistic stimuli to brain activity, particularly in fMRI studies of language comprehension \parencite{wehbe2014simultaneously, huth2016natural, deniz2019representation}. Most prior work fits encoding models at the voxel level, predicting each voxel’s response (independently) from stimulus-derived features \parencite{mitchell2008predicting, pereira2018toward,antonello_scaling_2024}. While voxelwise approaches offer fine spatial resolution, they rely on the implicit assumption that individual voxels constitute stable, functionally homogeneous units. In practice, however, individual voxels are noisy and often reflect a mixture of overlapping functional processes \parencite{naselaris2011encoding}. As a result, predictive information is fragmented across spatially correlated voxels, reducing statistical power and producing encoding patterns that are difficult to interpret in terms of coherent neural organization \parencite{huth2012continuous, deheer2017hierarchical}.

To mitigate the limitations of voxelwise modeling, several studies have explored encoding models defined over aggregated neural signals. Region-of-interest (ROI)–based approaches average responses across predefined anatomical or functional regions, improving signal-to-noise and interpretability \citep[e.g., ][]{ratan2021computational}. In the language domain, functional ROIs (fROIs) derived from independent localizer tasks have been widely used to study language-selective processing \parencite{fedorenko2010new, fedorenko2024language} and have been used to define subject-specific functional ROIs in encoding model studies \citep[e.g., ][]{schrimpf2021neural, alkhamissi-etal-2025-language}. While effective, these approaches require an explicit localizer task, which might not be available in all datasets, rely on pre-defined anatomical search spaces, which might not capture all language-relevant activity in an individual brain, and (by design) do not include language-evoked activity in other functional networks. Another approach for ROI-based analyses is to use anatomical masks that are fixed across participants \citep[e.g.,][]{tang2023semantic, oota2023joint}; although straightforward, this approach imposes fixed parcellations that may not align with subject-specific functional network structure.

Here, we build encoding models that predict brain responses not at the level of voxels or ROIs, but rather at the level of functional networks identified via independent component analysis (ICA). The motivation behind our approach is that the brain consists of multiple distributed, functionally coherent networks that can be identified from voxelwise activity timeseries\parencite{smith2009correspondence, yeo2011organization}. Given that these networks exhibit distinct functional specialization patterns, they can serve as meaningful units of analyses, complementing voxelwise and ROI-based approaches. 

Identifying functionally meaningful networks like the language network is a non-trivial problem without a dedicated localizer task. Traditional group-anatomical approaches rely on predefined anatomical boundaries or subject-averaged atlases, which often fail to recover functionally selective regions and cannot account for individual variability \citep[][]{Kanwisher_2010, fedorenko2010new}.  Individualized data-driven parcellation provides a solution: parcellation-derived cognitive networks align closely with networks identified with functional localizers \citep[][]{BRAGA2017457, du2024, Shain2025, salvo2025, Du_2025},  %for language and other higher-order networks (e.g., the multiple-demand and the default-mode networks), individualized data-driven network parcellations have highly similar topography to task-evoked responses, 
supporting the use of such parcellations to identify functionally coherent networks at the level of individuals, even in the absence of localizer tasks. In addition, ICA-based parcellation approaches are commonly used to denoise data, effectively separating stimulus-related neural signals from structured noise and artifacts \parencite{pruim2015icaaroma, griffanti2017hand}. ICA thus holds promise for encoding studies to simultaneously denoise and functionally parcellate brain activity into meaningfully distinct timecourses, without committing to the voxel as the fundamental unit of analysis.

%ICA has also been used to identify resting-state and task-related functional networks, demonstrating that large-scale brain activity can be decomposed into spatially distributed, temporally coherent components \parencite{beckmann2005investigations, smith2009correspondence, calhoun2001method}. Despite its widespread use, ICA is most often treated as a preprocessing or descriptive tool rather than as a representational substrate for predictive modeling.

Here, we propose an independent component–based encoding framework that shifts the focus from voxel-level prediction to modeling stimulus–response relationships at the level of latent functional networks. By training encoding models to predict component time series rather than individual voxel responses, this framework explicitly targets shared, network-level representations while attenuating voxel-specific noise (see \cite{skrill2024component} for a related approach in auditory neuroscience). We hypothesize that component-wise encoding models (i) effectively capture the variance in neural responses to naturalistic stimuli, resulting in high predictive performance, (ii) yield representations that are  interpretable in terms of cognitive and linguistic processing, and (iii) exhibit stability across runs and subjects, enabling prediction of the same funcional networks across individuals.

\begin{comment}
\subsubsection{Bridging encoding models and data-driven representations}
Recent work in cognitive computational neuroscience has increasingly emphasized the importance of choosing appropriate representational spaces when linking stimuli to neural responses \parencite{kriegeskorte2018cognitive}.  Building on this perspective, our work integrates ICA-derived functional components directly into the encoding framework, treating components as candidate representational units rather than intermediate denoising outputs. This approach complements existing voxelwise and ROI-based methods by offering a data-driven, network-level representational space that is both interpretable and amenable to cross-subject comparison.
\end{comment}
\section{Methods}

\begin{figure*}[ht]
  \begin{center}
   \includegraphics[width=\linewidth]{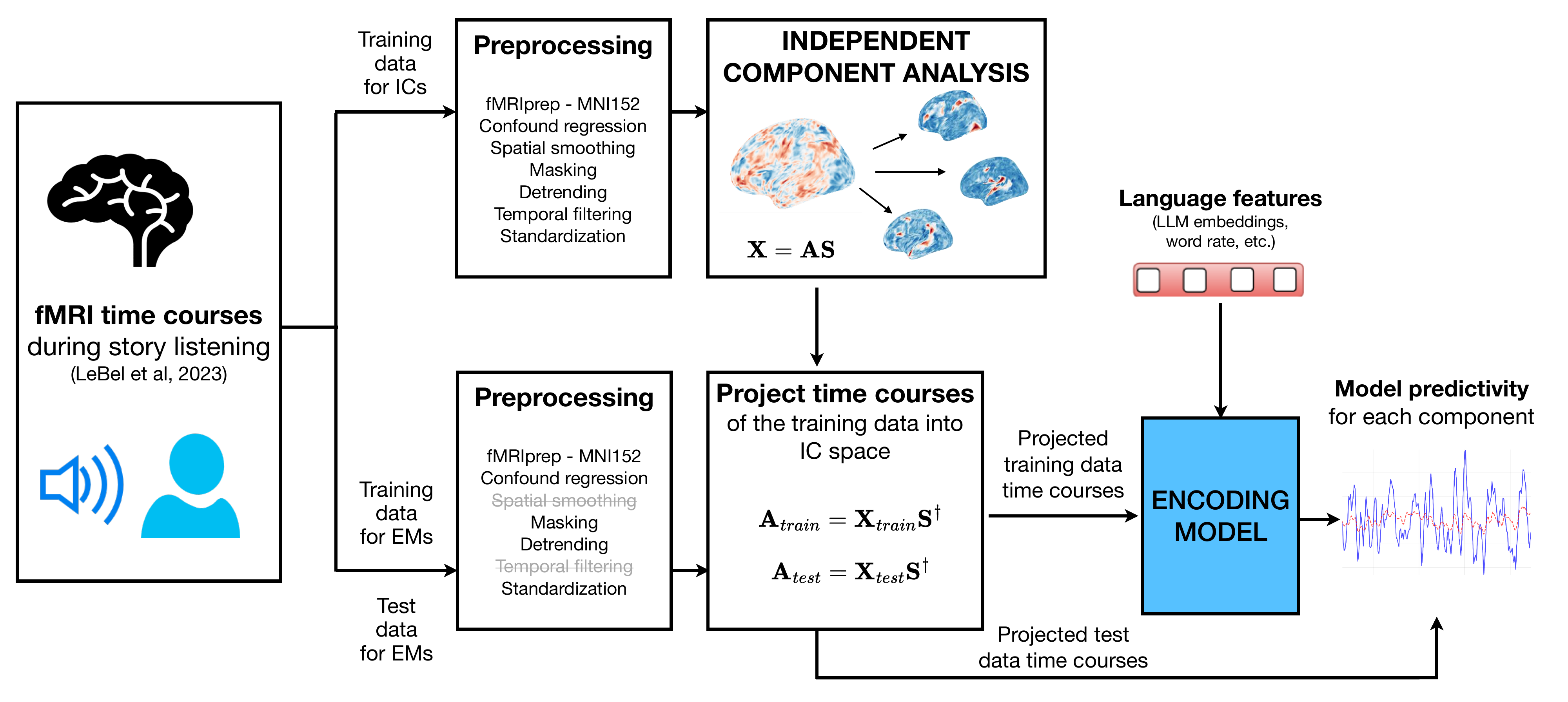}
    \end{center}
    \caption{Overview of the component-wise encoding model framework. fMRI data are preprocessed and decomposed using subject-specific spatial ICA trained on a subset of stories. The resulting spatial components are held fixed and used to project voxelwise data from the remaining stories into component space. Encoding models are trained to predict component time series from linguistic stimulus features and evaluated on a held-out test story.}
  \label{figure1}
\end{figure*}
\subsection{Data} 

We use the LeBel dataset \parencite{LeBel2023-bt}, a high quality fMRI dataset designed to support voxelwise encoding models and analyses of narrative comprehension. In this dataset, 8 healthy adult participants each listened to 26 complete natural narrative stories from \textit{The Moth Radio Hour}, with individual story durations of approximately 10–15 minutes. These stimuli amount to roughly 6 hours of rich, continuous spoken language per subject. To ensure independence between component estimation, model training, and performance evaluation, fMRI data for each subject were partitioned into three non-overlapping sets.
\begin{itemize}
    \item IC Estimation Set: Three stories ($\approx$30 min) reserved for defining the spatial components using ICA.
    \item Encoding Training Set: Twenty-two stories ($\approx$260 min) used to train the encoding models per component.
    \item Encoding Test Set: One story (\textit{Where There's Smoke}, $\approx$10 min) held out for testing. This story was repeated across multiple scan sessions and averaged to provide a high-Signal to noise ratio target for evaluating model predictivity.
\end{itemize}

\subsubsection{Preprocessing for IC Estimation Set:} fMRI data were first preprocessed using fMRIPrep \citep[][]{estebanFMRIPrepRobustPreprocessing2019}, including motion correction, slice-timing correction, susceptibility distortion correction, and normalization to MNI space, with all runs resampled to 2 mm isotropic resolution. Subsequent preprocessing for ICA estimation was performed using Nilearn’s NiftiMasker with a gray-matter mask. Time series were detrended and temporally band-pass filtered between 0.01 and 0.1 Hz. Confound regression was applied to remove motion-related and physiological noise, including the six rigid-body motion parameters, framewise displacement, first five aCompCor components, global signal, cosine drift regressors, and spike regressors for high-motion volumes (framewise displacement > 0.5). The data were then spatially smoothed with a 4 mm FWHM Gaussian kernel and standardized across time prior to ICA estimation.

\subsubsection{Preprocessing for Encoding Model Train and Test Set:} The same preprocessing pipeline was applied to the data used for training and testing the encoding model, including fMRIprep, gray-matter masking, detrending, confound regression and standardization. However, temporal band-pass filtering and spatial smoothing were not applied for these stories in order to preserve high-frequency temporal information and fine-grained spatial patterns relevant for model training. In addition, following the original protocol, the first and last 10 TRs were removed from each story to account for silence at the start of the scan as well as some portion of the story \citep[][]{LeBel2023-bt}. In addition, to account for long context artifacts, the first 50 TRs of the held-out test story were removed following \citep[][]{antonello_scaling_2024}.

\subsection{Independent Component Analysis} As mentioned previously, we use the IC Estimation Set to estimate our spatial components. This data is drawn from three different scanning sessions, allowing the decomposition to capture cross-session variability.

Let $\mathbf{X} \in \mathbb{R}^{T \times V}$ denote the preprocessed fMRI data matrix for a given subject, where $T$ is the number of time points and $V$ is the number of voxels. Spatial independent component analysis assumes that the observed voxelwise signals can be expressed as a linear mixture of statistically independent spatial components:
\begin{equation}
\mathbf{X} = \mathbf{A}\mathbf{S}
\end{equation}
where $\mathbf{S} \in \mathbb{R}^{K \times V}$ contains $K$ spatially independent component maps, and $\mathbf{A} \in \mathbb{R}^{T \times K}$ contains the corresponding component time series.
ICA estimates an unmixing matrix $\mathbf{W} \in \mathbb{R}^{K \times V}$ such that
\begin{equation}
\mathbf{S} = \mathbf{W}\mathbf{X}^\top
\end{equation}
with the rows of $\mathbf{S}$ being maximally statistically independent. The corresponding mixing matrix is given by $\mathbf{A} = (\mathbf{W}^{-1})^\top$.

After estimating the ICA decomposition on the training data, the spatial source matrix $\mathbf{S}$ was held fixed. For any new fMRI data matrix $\mathbf{X}_{\text{new}} \in \mathbb{R}^{T' \times V}$, component time series were obtained by projecting the voxelwise responses into the learned spatial basis:
\begin{equation}
    \mathbf{A}_{\text{new}} = \mathbf{X}_{\text{new}} \mathbf{S}^{\dagger}
\end{equation} where $\mathbf{S}^{\dagger}$ is the pseudoinverse of the component source matrix, yielding $\mathbf{A}_{\text{new}} \in \mathbb{R}^{T' \times K}$. These component time series served as targets for subsequent encoding model training.The number of components to be estimated were set to 100 per subject.

\subsubsection{ICA-AROMA} Following the initial ICA decomposition, ICA-AROMA (ICA-based Automatic Removal Of Motion Artifacts)\citep[][]{pruim2015icaaroma} was employed to identify and label noise-related components. This approach utilizes a pre-trained classifier to evaluate each independent component based on four robust spatial and temporal features: maximum fraction of high-frequency content, edge fraction, CSF fraction, and structural alignment. These components are not eliminated from the Sources, instead this is used for future analysis to understand encoding model predictivity.

\subsection{Encoding Models}
We trained encoding models to predict neural responses to naturalistic language stimuli from continuous linguistic features. Rather than modeling voxelwise responses directly, fMRI data were first projected into a lower-dimensional, data-driven representational space defined by ICs, allowing us to model shared and denoised neural signals across voxels. Encoding models were trained separately for each subject, component and was tested using the held-out story data.
\subsubsection{Linguistic features} We used a diverse set of linguistic predictors capturing both surface-level and high-level semantic information. These included word rate, lexical surprisal, and contextualized language model embeddings. Embeddings were extracted from the Pythia-410m model \parencite{biderman2023pythiasuiteanalyzinglarge}, to capture hierarchical representations of linguistic content across time. Pythia is a standard, open-source LLM that has been successfully used in prior studies to predict neural responses (e.g., \cite{alkhamissi-etal-2025-language}).
\subsubsection{Encoding model implementation} Voxelwise fMRI data were projected into ICA component space by applying the inverse transform of the learned spatial components, converting the data from shape (time points × voxels) to (time points × components). The resulting component time series served as the prediction targets for each subject. We used the LITcoder library to implement the encoding models \citep[][]{binhuraib2025litcoder}. Linguistic features were temporally aligned with the fMRI data using a finite impulse response model with delays spanning 5 TRs. Feature timeseries were downsampled to the fMRI acquisition rate using a Lanczos filter (window size = 3). Encoding models were trained using ridge regression with 5-fold cross-validation, with the regularization parameter selected to maximize generalization performance and mitigate overfitting. Prediction performance for each component was quantified as the Pearson correlation between the predicted and actual component timecourses in the test set. The full framework of the methods section is illustrated in Figure \ref{figure1}. We also compare IC EMs to traditional EM baselines (see Appendix, "Voxelwise and Anatomical ROI-based Encoding Models"), for baseline EM details.

\subsection{Identifying networks of interest}
Traditional voxelwise encoding analyses either examine the whole brain or define ROIs a priori, either using functional localizers or predefined anatomical or functional atlases. In contrast, our framework identifies putative functional regions in a fully data-driven manner by decomposing fMRI data into ICs, which are hypothesized to correspond to coherent neural sources underlying shared voxelwise activity patterns.

To relate these data-driven components to established brain regions, we adopted an atlas-matching procedure described by \citep[][]{Shain2025}. For each IC, spatial maps were first sign-corrected based on the median activation value to ensure consistent interpretation of positive loadings. We then thresholded each component by retaining only voxels within the top 99th percentile of absolute activation values, yielding spatially localized component maps.

The resulting thresholded IC maps were compared against reference parcellations from prior work \citep[][]{du2024, Lipkin2022-ca} (especially the Auditory (AUD), Language (LANG) and Visual (VIS) parcels) using spatial correlation. For each component, we computed correlations with parcels in the reference atlases and identified the parcel with the highest correspondence, allowing us to assign each IC to its closest known functional region. This procedure enabled direct comparison between data-driven components and canonical language/auditory/visual regions.

\subsection{Cross-subject component matching}
To assess which components reflect shared neural representations across individuals, we quantified component similarity across subjects using two complementary measures: temporal and spatial correlation. 

Temporal matching was performed by computing Pearson correlations between predicted component time series across subjects. For each component in a given subject, we identified the component in another subject whose time series exhibited the highest correlation, providing a measure of functional correspondence. After identifying the best temporal match, we computed the corresponding spatial correlation between the matched component maps.

Spatial similarity was assessed by computing spatial correlations between the ICA-derived component maps across subjects. Since all component maps were defined in MNI space, voxelwise spatial correlations could be computed directly, allowing us to quantify the degree to which components occupied similar anatomical locations across individuals.After identifying the best spatial match, we computed the corresponding temporal correlation between the matched component time series. Thus, in each case, one domain (temporal or spatial) was used for matching, and the other domain was used for evaluation.

\section{Results}
\begin{figure}[ht]
  \begin{center}
    \includegraphics[width=\linewidth]{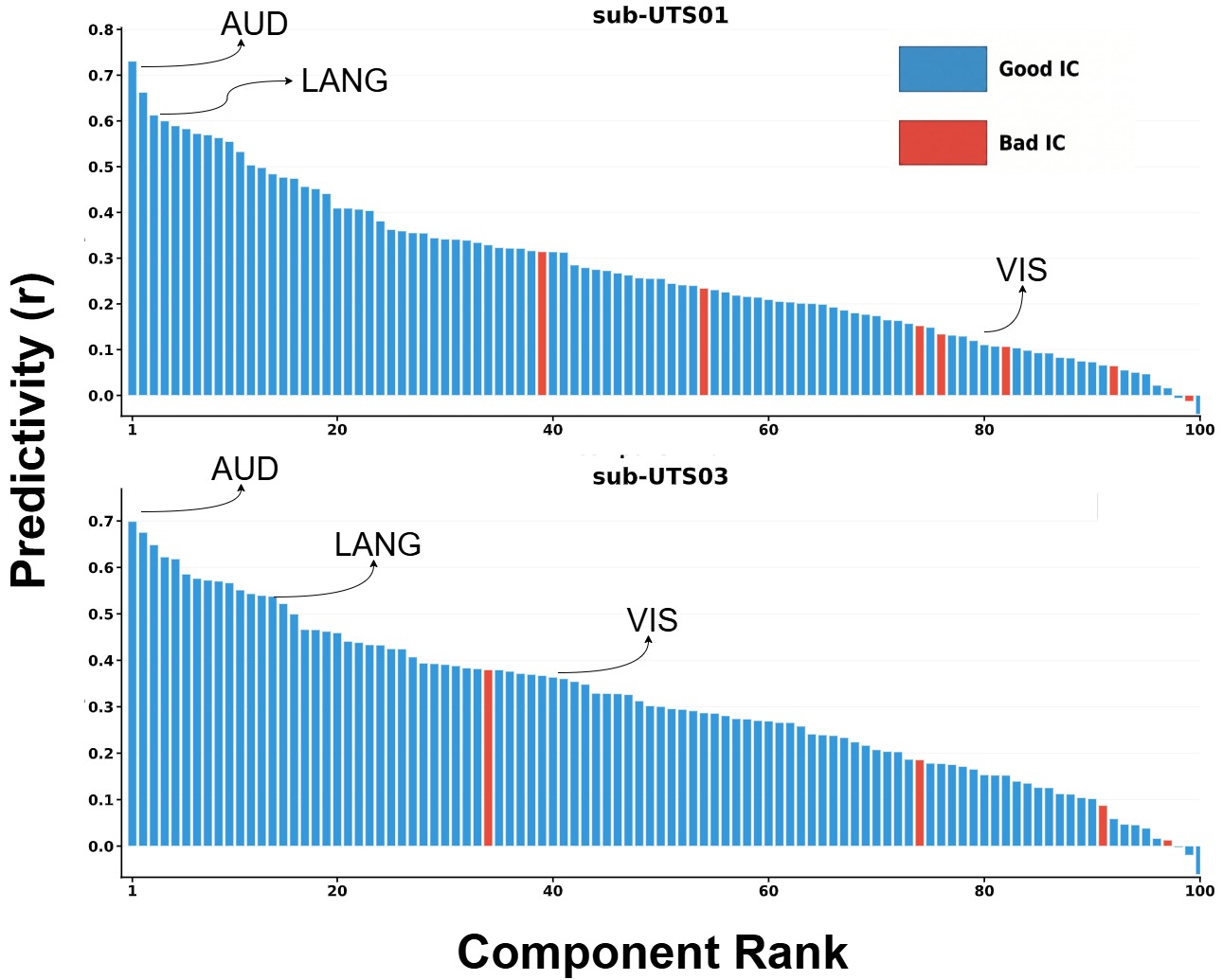}
  \end{center}
  \caption{Test story predictivity (correlation) across ICs for two sample subjects. ICs are ranked from best to worst. Blue bars denote "Good" ICs, which likely reflect meaningful neural activity; red bars represent "Bad" ICs (noise or artifacts), identified with AROMA. %The elbow graph shows that some components are better predicted than others.
  }
  \label{figure2}
\end{figure}

\begin{table}[!ht]
  \centering
  \caption{Summary of predictivity results across subjects showing the number of significant components after permutation testing and FDR correction.}
  \label{sample-table}
  \begin{tabular}{l p{1.8cm} p{1.8cm} p{1.8cm}}
    \hline
    \textbf{Subject} & 
    \textbf{Number of Sig. Components} & 
    \textbf{Mean~r (Sig. Components)} & 
    \textbf{Mean r} \\
    \hline
    Sub-UTS01 & 87 & 0.18 & 0.17 \\
    Sub-UTS02 & 92 & 0.17 & 0.16 \\
    Sub-UTS03 & 93 & 0.20 & 0.20 \\
    Sub-UTS04 & 92 & 0.14 & 0.13 \\
    Sub-UTS05 & 86 & 0.15 & 0.14 \\
    Sub-UTS06 & 89 & 0.11 & 0.10 \\
    Sub-UTS07 & 82 & 0.16 & 0.14 \\
    Sub-UTS08 & 89 & 0.10 & 0.09 \\
    \hline
  \end{tabular}
\end{table}
\begin{figure*}[!ht]
  \begin{center}
   \includegraphics[width=\linewidth]{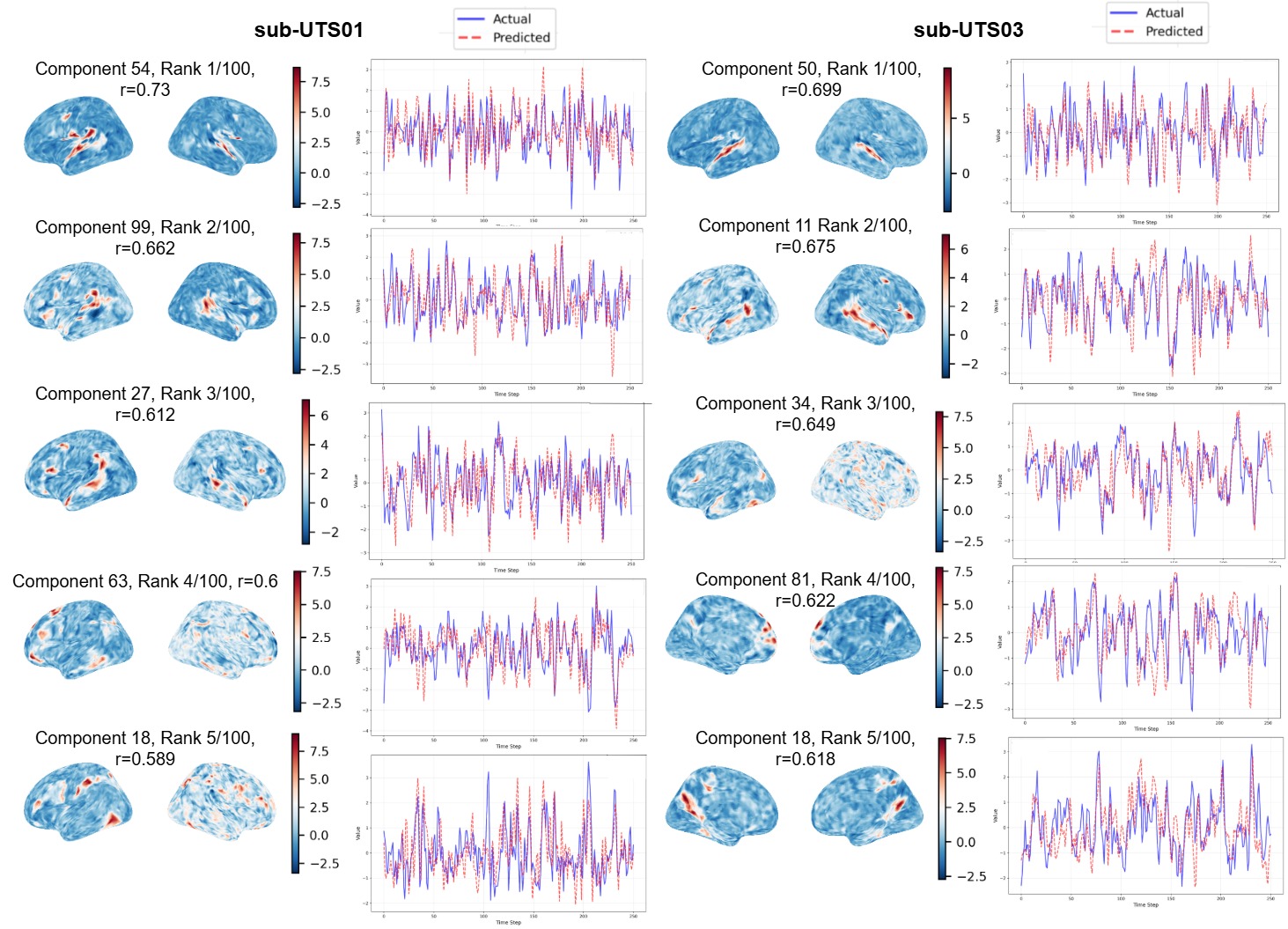}
    \end{center}
    \caption{ Top 5 most predicted ICs for representative subjects in the test dataset. Spatial weight maps (IC sources) and corresponding predicted vs. actual timecourses for the five highest-ranked components in subject sub-UTS01 (left) and sub-UTS03 (right). Predictivity is quantified via Pearson correlation ($r$) between the actual BOLD signal and the model-predicted response on held-out test data. This is a scaled figure; original in Appendix (Figure \ref{figureA4}).}
  \label{figure3}
\end{figure*}
\subsubsection{IC-based EM prediction reveals distinct predictivity across ICs}To assess the functional relevance of the decomposed neural signals, we examined the prediction correlation values on the test set across all ICs for each subject. As shown in Figure \ref{figure2}, predictive performance is highly non-uniform; a subset of components in each subject yielded high prediction correlations,with predictivity gradually decreasing across the remaining components.

To formally assess significance, we conducted permutation testing with the concatenated cross-validation folds. For each component and fold, stimulus features were temporally shuffled 1,000 times to generate a null distribution of prediction correlations. Multiple comparisons across components were controlled using False Discovery Rate (FDR) correction (q < 0.05). A large proportion of components survived permutation testing and FDR correction in every subject (82–93 ICs per subject; Table \ref{sample-table}). Mean prediction correlation across significant components ranged from r = 0.11 to r = 0.20.

To ensure that the high predictive power of certain components was not driven by motion or physiological artifacts, we utilized ICA-AROMA to classify components. As shown in Figure \ref{figure2}, the top components predicted by the encoding model do not consist of  noise as identified by ICA-AROMA (red bars). This confirms that highly predicted components reflect stimulus-driven neural signals rather than noise. 

Figure \ref{figure3} shows the top 5 predicted components for sub-UTS01 and sub-UTS03 on the held-out test set. These components demonstrate strong temporal alignment between predicted and observed time series, with prediction correlations exceeding r$\approx$ 0.60 for the highest ranked components. Notably, top components often corresponded to auditory or language networks.

\begin{figure*}[!ht]
  \begin{center}
   \includegraphics[width=\linewidth]{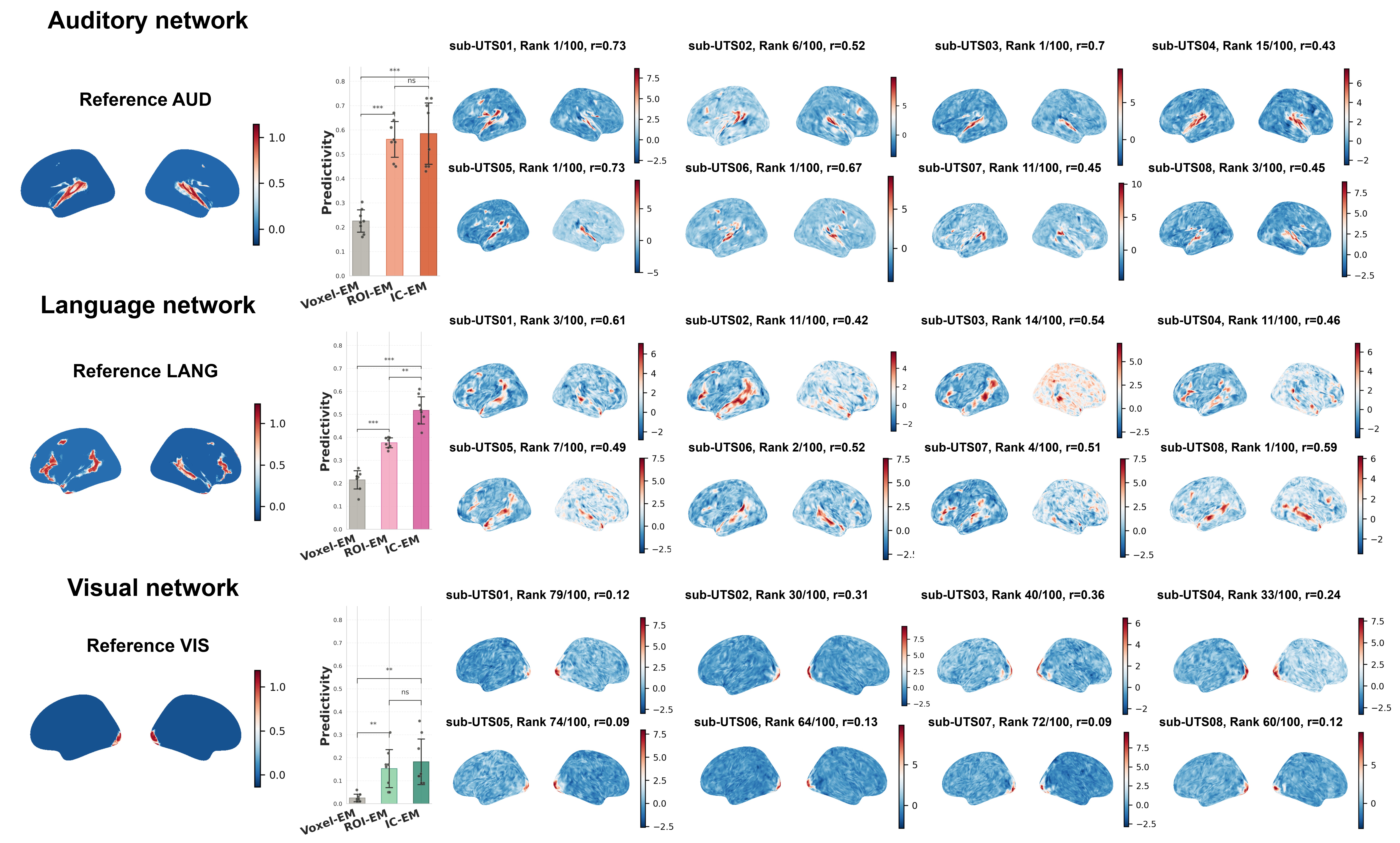}
    \end{center}
    \caption{Predictivity of Auditory, Language and Visual components across subjects. Functional networks are chosen from ICs by computing the spatial correlation between the ICs and the reference atlases shown in the figure. The IC with the highest spatial correlation to the reference is labeled as the corresponding functional network for the subject. Bar height represents the mean predictivity (Pearson r) across subjects (n = 8) for each encoding model type: Voxel-EM, ROI-EM  and IC-EM. Error bars denote ±1 standard deviation.  Pairwise paired t-tests with Benjamini–Hochberg FDR correction: * p < 0.05, ** p < 0.01, *** p < 0.001; ns = not significant.To the right, subject specific functional networks identified through the above procedure are shown with their corresponding rank and predictivity scores. The auditory component exhibits the strongest and most consistent predictivity across subjects. The Language network also exhibits good predictivity across subjects, sometimes coming in right next to the auditory network or even the most predictive (for sub-UTS08). In contrast, the visual network shows weak predictivity consistent with the task at hand (listening to naturalistic stimuli).}
  \label{figure4}
\end{figure*}

\vspace{8pt}
\subsubsection{Auditory, Language, and High-Order Cognitive networks have high predictivity } To evaluate network-specific encoding model performance, we examined predictivity across the Auditory, Language and Visual network for all subjects. Using the procedure mentioned in the Methods sections to identify regions of interest, we identify the AUD, LANG and VIS network for each subject from the ICs. From Figure \ref{figure4}, we can see that the auditory network showed the strongest and most consistent predictivity across subjects. Group-level performance was high (mean r $\approx$ 0.59). %Spatial maps also demonstrate a clear alignment to the reference and between subjects.
The language network demonstrated moderate predictivity (mean r $\approx$ 0.52). While rankings were generally slightly lower than the auditory network, language components consistently appeared within the top 15 across subjects. %Spatially too, they are moderately aligned to the reference atlas.
In contrast, the visual network exhibited weak predictivity (mean r $\approx$ 0.18) and low component rankings across subjects, often falling in the bottom half of components. This reduced performance is consistent with the auditory nature of the task (naturalistic story listening), which imposes limited visual processing demands.

In addition, we also have a look at other higher order cognitive networks like FPN\_A, FPN\_B, DN\_A and DN\_B (from \cite{du2024}) in the Appendix section --Analysis on additional known functional networks. From Figure \ref{figureA5} we can see that these networks are also well predicted for this stimuli. 

Compared to IC EMs, voxelwise EMs (for voxels within the reference atlas) show far weaker predictivity (Figure \ref{figure4}). 
%\cor{From Figure \ref{figure4} we can observe that compared to the IC EM, voxelwise EMs have lesser predictivity. 
Anatomical ROI-based EMs (predicting the average timeseries within the reference atlas) perform on par with IC EMs (t-test not significant) for the AUD and VIS networks (which have a relatively fixed anatomical position), but  the language network---which is known to be anatomically variable across subjects \citep[][]{fedorenko2024language}---is better predicted by individualized IC-based EMs.

\begin{figure}[ht]
  \begin{center}
    \includegraphics[width=\linewidth]{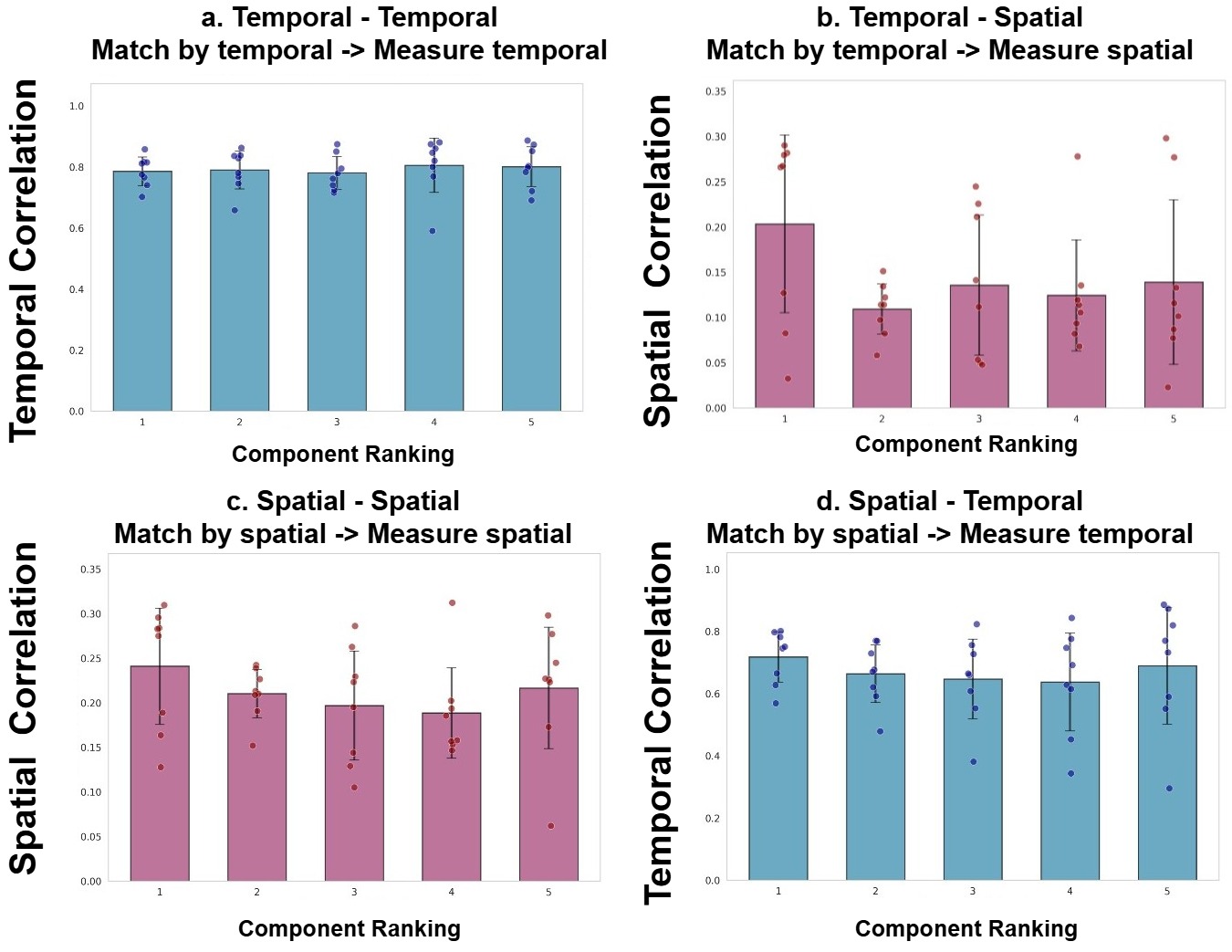}
  \end{center}
  \caption{Temporal vs.~spatial component matching strategies. Top row (Temporal-first [a,b]): Components matched by maximizing temporal correlation, showing strong temporal agreement (left) and moderate spatial agreement (right) of matched components. Bottom row (Spatial-first [c,d]): Components matched by maximizing spatial correlation, showing strong spatial agreement (left) and moderate temporal agreement (right) of matched components. Bar heights represent mean correlations across all subjects (n=8), error bars show standard deviations, and individual dots represent subject-level values. Both matching strategies achieve robust cross-subject component correspondence, with each approach excelling in its respective matching domain while maintaining reasonable performance in the complementary domain.}
  \label{figure5}
\end{figure} 

\subsubsection{Component Matching across subjects} Cross subject component matching is done using spatial and temporal correlation. To get the group level estimates, we deployed a leave-one-out approach, where one subject was selected as a reference, and for the remaining subjects, components were identified according to the matching criterion (temporal or spatial). The evaluation metric (again spatial or temporal) is computed, and correlations were then averaged across all non-reference subjects, providing a mean for each subject. This procedure is then repeated with each subject serving as a reference. The final reported value for each component rank reflects the average of these subject-specific mean correlations as shown in Figure \ref{figure5}.  As observed in Figure \ref{figure5}, for the top five components across all subjects, both matching strategies yielded comparable correspondence patterns. When matching components temporally and evaluating spatial similarity, spatial correlations were consistently positive, indicating that functionally aligned components exhibit meaningful spatial similarity across individuals. Conversely, when matching components spatially and evaluating temporal similarity, temporal correlations were robust, demonstrating that spatially aligned components preserve shared functional dynamics. Together, these findings indicate that the highly predicted components exhibit cross-subject consistency in both temporal dynamics and spatial organization. While absolute spatial correlations are modest as expected given inter-individual variability in cortical topography, both matching strategies successfully recover shared functional structure across participants.

\begin{figure}[ht]
  \begin{center}
    \includegraphics[width=\linewidth]{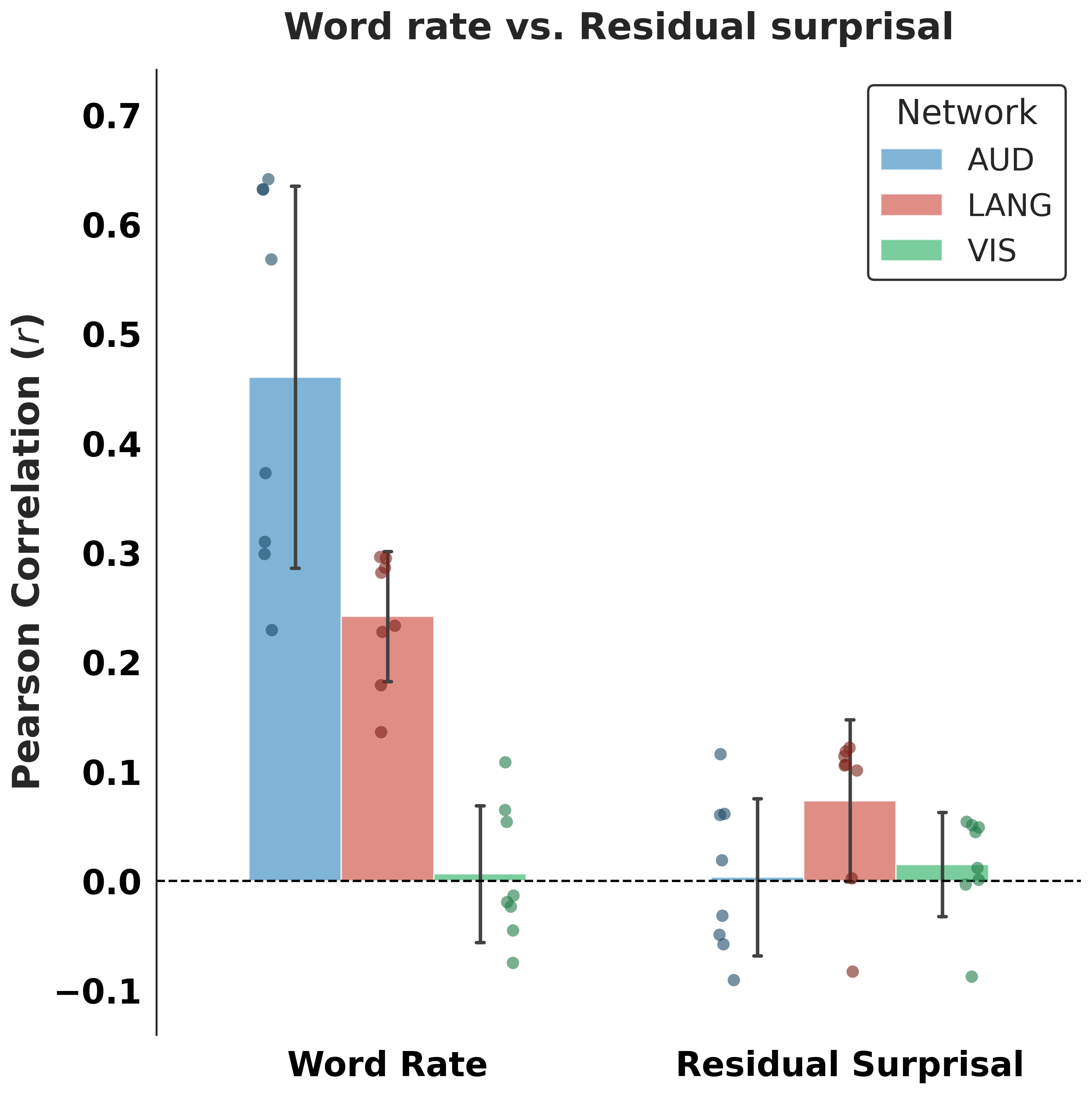}
  \end{center}
  \caption{Word rate and residual surprisal predictivity across brain networks. Bars represent mean correlations with standard deviation error bars. Dots represent individual subjects. As expected, word rate is a substantial predictor for AUD and moderate predictor for LANG. Residual surprisal is a bad predictor of AUD but moderate predictor for LANG. VIS is used as a control as this task does not involve visual stimuli.}
  \label{figure6}
\end{figure}

\subsubsection{Word rate and Surprisal Effects on Network Responses}
To characterize low- and higher-level linguistic contributions to network activity during naturalistic listening, we examined the effects of word rate and surprisal.

Word rate is defined as the number of words per unit time (TRs), therefore capturing low-level speech timing and acoustic-structural density. Word rate primarily reflects fluctuations in speech input rate and is closely related to auditory processing demands.

Surprisal is defined as the negative log probability of a word given its preceding context:
\begin{equation}
    S(w) = -\log P(w | \text{context})
\end{equation}
Surprisal quantifies how unexpected a word is in context and reflects higher-level predictive language processing demands.

Word rate and surprisal are positively correlated, as faster speech segments tend to contain more unpredictable lexical content. To dissociate low-level input density effects from higher-level linguistic prediction effects, we regressed word rate out of surprisal and used the residualized surprisal time series in subsequent analyses.
%As observed in Figure \ref{figure6} encoding model predictivity on test-set revealed that 
Word rate strongly predicted activity in auditory cortex, consistent with sensitivity to low-level acoustic and temporal speech structure (Figure \ref{figure6}). Word rate also showed moderate predictive power within the language network.
In contrast, the visual cortex (our control region) showed minimal prediction by either regressor, as expected given the auditory nature of the task.

After regressing out word rate, residual surprisal continued to predict activity within the language network more than in the auditory network. While this effect is relatively modest, it suggests that language-selective regions are sensitive not only to speech rate or acoustic density, but also to aspects of higher-order predictive structure in linguistic input.
More broadly, these findings are consistent with accounts of hierarchical stimulus tracking during naturalistic listening, in which early auditory regions primarily track input density and temporal speech structure, while downstream language-selective regions may additionally reflect higher-level predictive information \citep[][]{Caucheteux_Gramfort_King_2023}.

Importantly, this analysis highlights the utility of the component-based encoding framework. By examining feature selectivity at the network level, these IC encoding models provide a testbed for interpretability analyses, allowing dissociation of low-level acoustic drivers from higher-order linguistic computations.

\section{Discussion}
This work introduces an IC–based encoding framework that shifts the unit of analysis from individual voxels to data-driven functional components. By training encoding models on component time series, we aim to capture stimulus–brain relationships at a representational level that better reflects the distributed nature of cortical processing during naturalistic language comprehension.
At the same time, the approach retains the core design features of brain encoding models: flexible feature representations (e.g., derived from AI models) and linking functions (in our case, linear regression).

A key implication of our work is that independent components provide a meaningful representational basis for encoding models. Previous work has indicated that ICA-derived components capture coherent patterns of activity that reflect the underlying functional organization \citep[][]{du2024, Shain2025}. We show that IC-based encoding models also yield interpretable results, e.g., language and auditory networks are consistently well predicted during our naturalistic listening paradigm, whereas the early visual network shows substantially lower predictivity (Figure \ref{figure4}). Components that yield strong encoding performance show substantial spatial and functional consistency across subjects (Figure \ref{figure5}); this convergence suggests that predictive components correspond to shared functional networks rather than subject-specific or artifactual patterns. Further, component-wise encoding facilitates cross-subject comparison without relying on predefined atlases, while retaining interpretability at the network level. 
%remains compatible with standard encoding pipelines %\parencite{binhuraib2025litcoder} and feature representations.

Analyses on word rate and surprisal further support a hierarchical organization of component responses. Low-level features such as word rate have high predictivity for the auditory network, but account for relatively limited variance as compared to higher-level language embeddings in the language network (Figure \ref{figure6}). This pattern suggests that while acoustic and temporal features capture early sensory processing, richer semantic representations are required to explain activity in higher-order language and semantic networks.

The proposed framework provides a useful basis for future analyses. By providing a shared component space that preserves inter-subject variability, it enables systematic testing of hypotheses about how different linguistic features (e.g., semantic, syntactic, or acoustic representations) map onto brain activity during naturalistic language comprehension (or across various tasks). In addition, the framework provides a principled basis for studying cross-subject generalization, allowing models to transfer across individuals while retaining subject-specific structure. This creates opportunities to investigate which aspects of neural responses are shared versus idiosyncratic, and how these vary across individuals or experimental contexts.
As with any data-driven decomposition approach, certain methodological factors may influence the resulting components.ICA decomposition is inherently influenced by model order (i.e., the number of independent components specified). In the present study, preliminary analyses across different IC numbers yielded qualitatively similar results, suggesting that the main findings are reasonably robust to this parameter. Nonetheless, a more systematic evaluation of model order selection could further clarify its influence and provide additional validation in future work.

Overall, this study demonstrates that independent components provide a stable, interpretable, and functionally meaningful substrate for neural encoding models, offering a complementary approach to voxelwise approaches for mapping between AI models and the human brain.

\section{Acknowledgments and Disclosures}
{We thank the members of the LIT lab for their valuable feedback and discussions that helped improve this work. Claude (Anthropic) was used to reformat code (with author verification) and draft GitHub documentation.}

\FloatBarrier

\printbibliography

\section{Appendix}
\subsection{Cross-validation Results}
\begin{figure}[ht]
  \begin{center}
    \includegraphics[width=\linewidth]{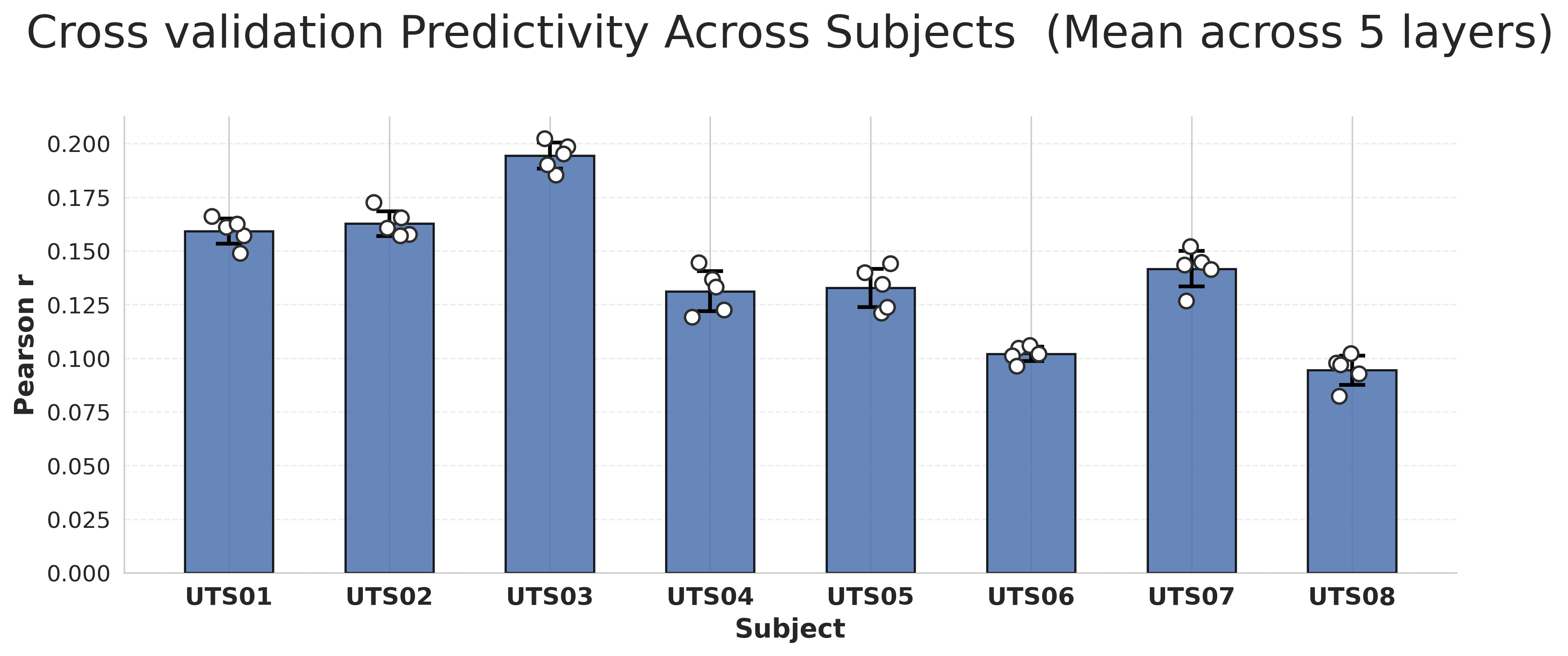}
  \end{center}
  \caption{Mean predictivity across 5-fold cross-validation, across all subjects. Results show that the observed patterns are consistent across different train/test splits, indicating that the findings are not driven by idiosyncratic properties of a single held-out story. Error bars represent ±1 standard deviation across folds.}
  \label{figureA1}
\end{figure}

\begin{figure}[ht]
  \begin{center}
    \includegraphics[width=\linewidth]{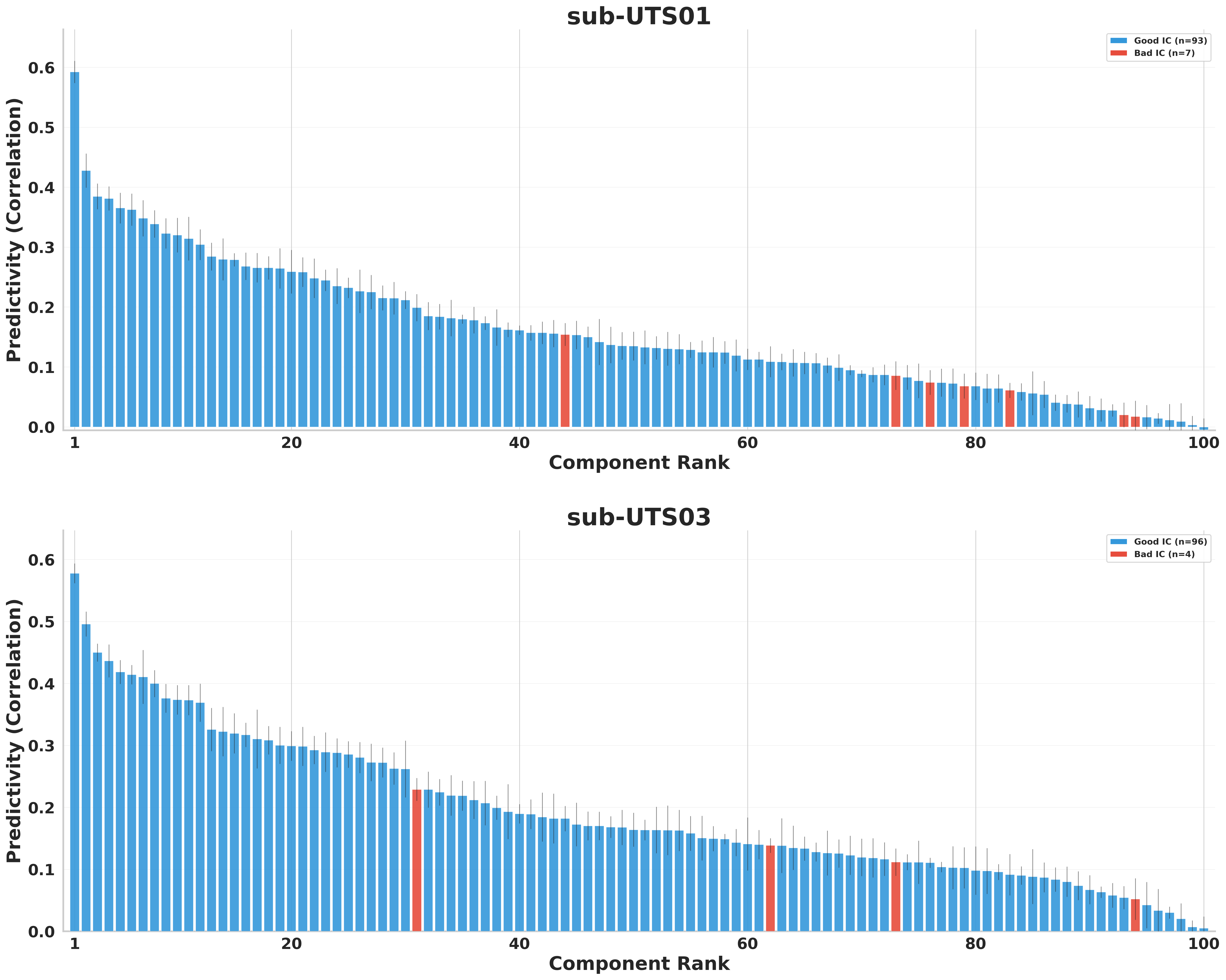}
  \end{center}
  \caption{Cross-validation predictivity of components for participants UTS01 (top) and UTS03 (bottom). Components are ranked in descending order based on the predictivity of the component along the x-axis, with the y-axis showing mean predictivity. Error bars indicate ±1 standard deviation across 5-fold cross-validation, illustrating the stability of each component’s contribution across splits. As we can observe, the patterns for both subjects are similar to Figure 2. One speculative reason as to why Figure 2 has higher predictivity compared to this might be because the data for the test story is cleaner as it is averaged across multiple runs for each subject.}
  \label{figureA2}
\end{figure}

To assess the stability of our results across train/test splits, we performed 5-fold cross-validation and summarize the outcomes in Figures \ref{figureA1} and \ref{figureA2}. Figure \ref{figureA1} demonstrates that subject-level predictivity remains highly consistent across folds, indicating that our findings are not driven by a particular held-out story. Figure \ref{figureA2} further shows that component-level rankings are similarly stable for individual participants, with error bars reflecting minor variance across folds. Notably, thecross-validated predictivity values are slightly lower than those in the main-text analyses; this is expected given that the test story used in the main analysis is averaged over multiple repetitions and thus exhibits a higher SNR. Together, these results confirm that both voxel- and component-level effects are robust to differences in story partitions and are not dependent on a single test set.

\subsection{voxelwise and ROI-based Encoding Models}
To contextualize the proposed IC-based framework, we implemented both voxelwise and anatomical ROI-based encoding models as baselines.

\subsubsection{voxelwise encoding models}
For each voxel, we trained an independent encoding model to predict its fMRI time course from the stimulus features. Model performance was evaluated using Pearson correlation between predicted and actual time series, averaged across cross-validation folds. For ROI-level comparisons, voxelwise prediction scores were aggregated by applying an anatomical atlas and computing the mean predictivity across all voxels within each ROI.

\vspace{10pt}

\subsubsection{ROI-based encoding models}
In the ROI-based approach, we first averaged the fMRI time series across all voxels within each anatomical ROI to obtain a single representative signal per region. The ROIs considered here are the AUD and VIS atlases from \citep[][]{du2024} and the Language network from \citep[][]{Lipkin2022-ca} (Reference images shown in Figure 3). We then trained encoding models to predict these ROI-averaged time series directly from the stimulus features. Performance was again quantified using Pearson correlation between predicted and actual ROI signals.

\subsection{Analysis on additional known functional networks}
\begin{figure*}[ht]
  \begin{center}
   \includegraphics[width=\linewidth]{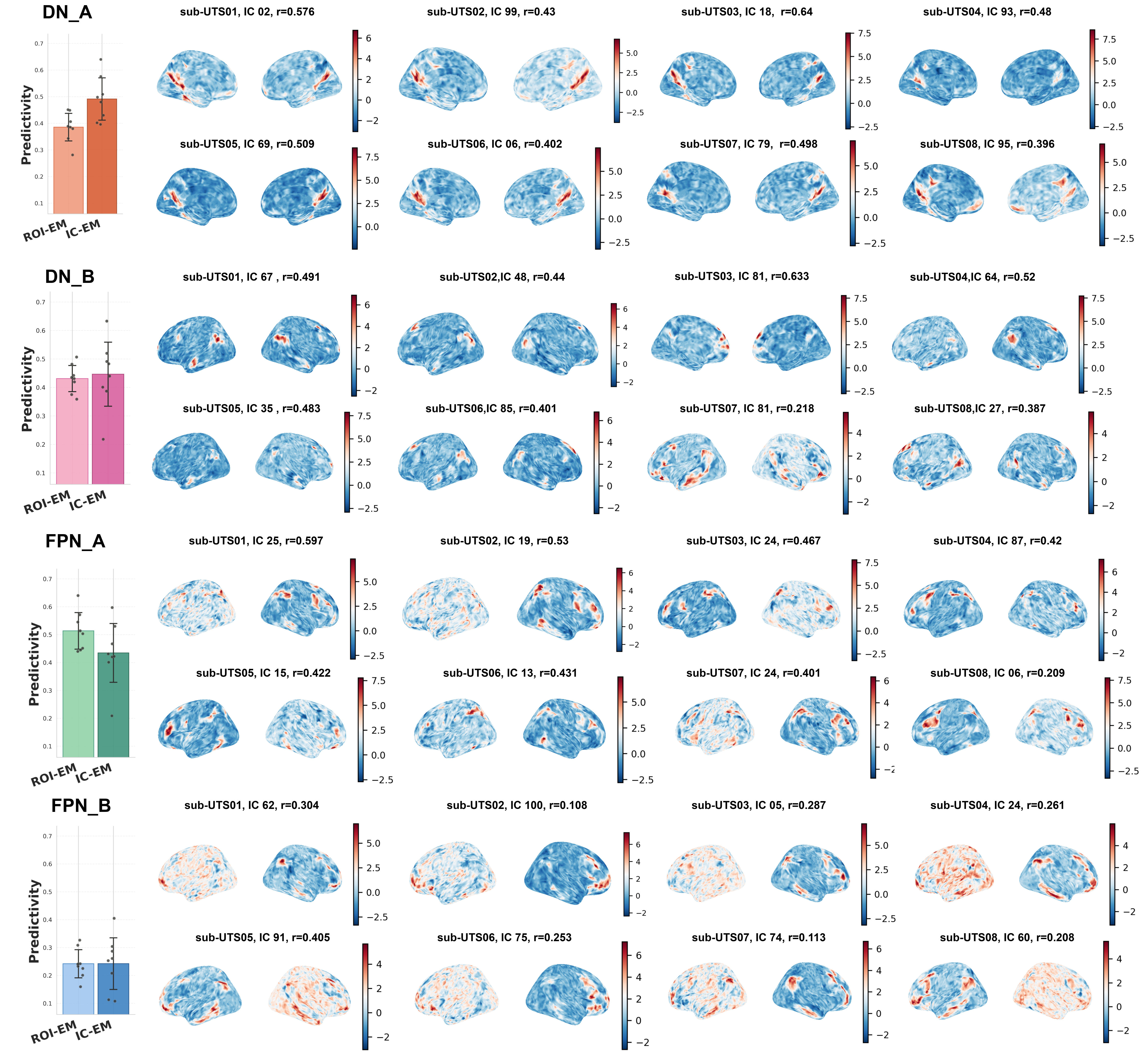}
    \end{center}
    \caption{Predictivity scores of networks most spatially correlated to the networks found in DU atlases for each subject. Networks such as DN\_A, DN\_B, FPN\_A are highly predictive across subjects, showing that there might be engagement of these networks during the task of listening to naturalistic stimuli.}
  \label{figureA5}
\end{figure*}

Beyond the language network, we observe that several additional large-scale functional networks from \citep[][]{du2024}, particularly subdivisions of the frontoparietal (FPN\_A, FPN\_B) and default mode (DN\_A, DN\_B) networks that are also robustly predicted across subjects (Figure \ref{figureA5}). Notably, these networks exhibit high predicitvity, potentially reflecting higher-order integrative and contextual processing during story comprehension. In this sense, the IC-based framework does not merely improve prediction performance, but provides a principled, data-driven means of identifying and dissociating the contributions of multiple interacting brain networks, thereby enabling more targeted and interpretable scientific investigation beyond what is typically done with standard ROI- or voxel-based analyses.

\begin{figure*}[ht]
  \begin{center}
   \includegraphics[width=\linewidth]{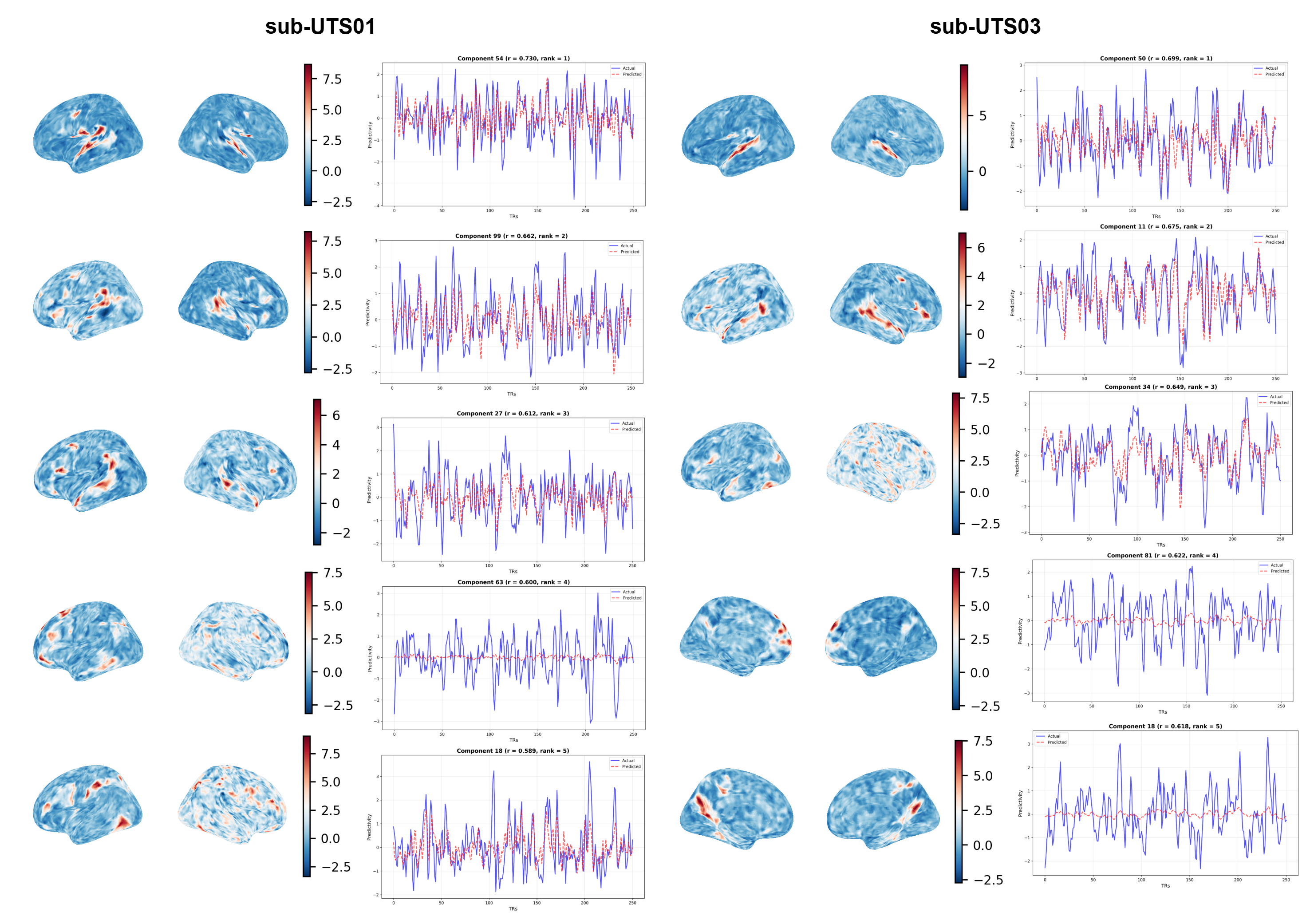}
    \end{center}
    \caption{ Top 5 most predicted ICs for representative subjects in the test dataset. Spatial weight maps (IC sources) and corresponding predicted vs. actual timecourses for the five highest-ranked components in subject sub-UTS01 (left) and sub-UTS03 (right). Predictivity is quantified via Pearson correlation ($r$) between the actual BOLD signal and the model-predicted response on held-out test data.We observe that some independent components (ICs) exhibit high Pearson correlation between predicted and actual time series despite clear differences in signal amplitude. This reflects a known property of correlation as a metric: it is invariant to linear scaling and therefore primarily captures similarity in temporal dynamics rather than absolute magnitude. As a result, models can accurately track the shape and timing of fluctuations while underestimating or overestimating their amplitude.Several factors may contribute to this dissociation. First, preprocessing steps such as run-wise normalization or standardization can attenuate meaningful amplitude differences across time, making it more difficult for models to recover absolute signal scale. Second, regularization in the encoding model may bias predictions toward smaller magnitudes, particularly in components with lower signal-to-noise ratio. Third, different ICs may vary in how strongly their underlying neural activity is reflected in the measured BOLD signal, leading to variability in how well amplitude can be captured even when temporal structure is preserved. Future work would investigate this further.}
  \label{figureA4}
\end{figure*}

\end{document}